\documentclass[sigconf]{acmart}
\usepackage[utf8]{inputenc}

\usepackage{amsmath,amssymb,amsthm}
\usepackage{algorithm}
\usepackage{algorithmic}
\usepackage{caption}
\usepackage{ltablex}
\usepackage{booktabs}
\usepackage{hyperref}
\usepackage{subcaption}
\usepackage{url}
\usepackage{color}
\usepackage{graphicx}
\usepackage{xcolor}

\usepackage{enumitem}

\usepackage{array}
\usepackage{multicol}

\usepackage{caption}
\usepackage{booktabs}
\usepackage{float}
\usepackage{etoolbox}

\usepackage[bottom]{footmisc}
\usepackage{xcolor}
\usepackage{booktabs}

\usepackage{array}
\newcolumntype{L}[1]{>{\raggedright\let\newline\\\arraybackslash\hspace{0pt}}m{#1}}
\newcolumntype{C}[1]{>{\centering\let\newline\\\arraybackslash\hspace{0pt}}m{#1}}
\newcolumntype{R}[1]{>{\raggedleft\let\newline\\\arraybackslash\hspace{0pt}}m{#1}}
\newcommand{\bx}{\ensuremath{\mathbf{x}}}
\newcommand{\bX}{\ensuremath{\mathbf{X}}}
\newcommand{\bz}{\ensuremath{\mathbf{z}}}

\DeclareMathOperator*{\argmin}{argmin}
\newtheorem{assumption}{Assumption}

\definecolor{dgreen}{RGB}{63, 175, 115}
\acmDOI{}

\acmISBN{}

\acmConference[]{}{}{}
\acmYear{2017}
\copyrightyear{2017}

\acmPrice{}

\begin{document}
\sloppy

\title{Prophit: Causal inverse classification for multiple continuously valued treatment policies}
\author{Michael T.~Lash}

\affiliation{%
\institution{University of Iowa}
\department{Department of Computer Science}
\city{Iowa City}
\state{IA}
\postcode{55242}		
}
\email{michael-lash@uiowa.edu}

\author{Qihang Lin}
\affiliation{%
\institution{University of Iowa}
\department{Management Science Department}		
\city{Iowa City}
\state{IA}
\postcode{55242}
	}
\email{qihang-lin@uiowa.edu}
	
\author{W.~Nick Street}
\affiliation{%
\institution{University of Iowa}
\department{Management Science Department}		
\city{Iowa City}
\state{IA}
\postcode{55242}
	}
\email{nick-street@uiowa.edu}

\begin{abstract}
Inverse classification uses an induced classifier as a queryable oracle to guide test instances towards a preferred posterior class label. The result produced from the process is a set of instance-specific feature perturbations, or recommendations, that optimally improve the probability of the class label. In this work, we adopt a causal approach to inverse classification, eliciting treatment policies (i.e., feature perturbations) for models induced with causal properties. In so doing, we solve a long-standing problem of eliciting multiple, continuously valued treatment policies, using an updated framework and corresponding set of assumptions, which we term the inverse classification potential outcomes framework (ICPOF), along with a new measure, referred to as the individual future estimated effects ($i$FEE). We also develop the approximate propensity score (APS), based on Gaussian processes, to weight treatments, much like the inverse propensity score weighting used in past works. We demonstrate the viability of our methods on student performance.
\end{abstract}

\keywords{Inverse classification, Causal learning, Propensity score, Multiple treatments, Continuous treatments, Deep learning}

\maketitle

\section{Introduction}


In a typical inverse classification setting, the goal is to elicit instance-specific feature perturbations that optimally direct the instance towards a preferred classification. In such a setting a classifier is used as a queryable oracle, in conjunction with the instance's feature vector, to elicit such recommendations (i.e.,~feature value perturbations).
The inverse classification process takes the form of an optimization problem, where the real-valued output of the classification function (e.g.,~probabilities elicited from a logistic regression model) is used as the objective function that is minimized or maximized, depending upon the class of interest (e.g.,~the probability of a ``bad'' classification is minimized). Furthermore, constraints imposed on the optimization process, along with several other considerations, further encourage the inverse classification process to produce real-world operational recommendations.

Considerations have not been made to address the inherent causal nature of the problem being solved, however. In short, the recommendations elicited from the inverse classification process can be viewed as a \textit{treatment policy} that produces a desired \textit{effect}, where the desired effect is maximal improvement in the probability (or some notion thereof) of a desired classification. This quantity -- the amount of outcome improvement gained by taking the treatment(s) -- is referred to as the \textit{individual treatment effect} (ITE) in causal inference literature and is precisely the quantity inverse classification is seeking to maximize.

A causal interpretation of (and by extension, causal methodology applied to) the inverse classification problem is intuitive: consider, for example, a scenario wherein a classifier has been induced to learn the mapping from student lifestyle feature vectors to end-of-term grade outcome (as in some of our experiments). In this scenario, when a new [test] student is encountered, an initial probability estimate of incurring a bad grade is obtained from the induced classifier. Then, inverse classification is performed by working backwards through the classifier, which produces a set of lifestyle perturbations, representing the changes the instance would need to make in order to optimally reduce their probability of a bad grade. In other words, changing ones lifestyle \textit{causes} a reduction in bad grade probability -- the effect! As can be seen through this simple example, the problem is inherently causal in nature, and thus requiring causal adaptions to the inverse classification framework (explained in Section 2).

To such an end, we propose \textcolor{red}{Pro}\textcolor{blue}{phit}, a neural network-based classifier and adapted inverse classification framework that elicits causal inverse classification recommendations. The causal design of our classifier and framework centers around fitting (\textcolor{blue}{phit}) an \textit{approximate} \textcolor{red}{pro}pensity score-weighted (APSW) classifier (hence, \textit{Prophit}), which is obtained using Gaussian processes (GPs). The ensuing Prophit-based framework is subsequently adapted to trade off optimizing for causality and probabilistic improvement by using a regularizer in conjunction with a maximum likelihood estimation approach to causal feature optimization. This method and framework are described in Section 2.

With a brief overview of Prophit in mind, we enumerate our contributions as follows:
\begin{enumerate}
    \item We augment an existing inverse classification framework with causal properties to elicit causal, individual-specific treatment policies.
    \item We create a Gaussian process-based \textit{approximate propensity score} (APS) that is capable of approximating treatment propensity in a flexible, nonlinear manner for handling multiple, continuously valued treatments.
    \item Using the APS, we create causal classifiers that are an elaboration on the current state of the art.
    \item We develop $i$FEE (individual future estimated effects), a measure that is an elaboration on ITE, along with an update on the potential outcomes framework (POF), which we term ICPOF (the inverse classification potential outcomes framework), and an updated set of assumptions, to account for the inverse classification component of the problem and the mutltiple, continuously valued treatments.
    \item Our propositions regarding the use of APS-reweighted classifiers and the ensuing causal inverse classification framework solve the long-standing causal problem of obtaining optimal treatment policies using multiple, continuously-valued treatments (in accord with our assumptions).
\end{enumerate}

The rest of the paper proceeds with a disclosure of our methodology in Section 2, followed by our experiments and experiment methodology in Section 3, and related work in Section 4. Section 5 concludes the paper.

\section{Causal Inverse Classification}

In this section, we relate our methods of making the inverse classification process elicit causal treatment policies and, furthermore, how we adapt an existing inverse classification framework to trade-off optimizing for causality and probabilistic improvement in outcome.

\subsection{Preliminaries}

Prior to discussing our methods, we begin by defining some preliminary notation that will be used throughout the discourse of this work.

Let $\{(\bx^{(i)}_{C},\bx^{(i)}_{I},\bx^{(i)}_{T},y^{(i)})\}_{i=1}^{n}$ denote a dataset of $n$ instances, where $\bx^{(i)} \in \mathbb{R}^{p}$ is instance $i$'s full feature vector and where $C$, $I$, and $T$ are index sets that refer to specific subvectors of $\bx^{(i)}$; $y^{(i)} \in \{0,1\}$ is the event label.

The index sets $C$, $I$, and $T$ reference what are referred to as the ``control'', ``indirectly changeable'', and ``treatment'' features, respectively, and are defined to afford certain considerations with respect to the inverse classification process\footnote{Note that in several past inverse classification works \cite{Lash2017a,Lash2017b}, these were referred to by $U$,$I$,$D$, representing the ``unchangeable'' (i.e.,~control), ``indirectly changeable'' (unchanged in this work), and ``directly changeable'' (i.e., treatment)}. More specifically, by distinguishing between these respective subvectors, we can optimize over only those features that can actually be changed, or that the inverse classification designer want to be used as treatments; in our running student performance example these are lifestyle features such as time spent with friends. The control and indirectly changeable feature sets will be discussed in greater depth shortly.

In this work, we adopt the inverse classification framework of Lash et al.~ \citep{Lash2017b}, which we will modify to suit our causal endeavors, and which we have updated to reflect our specified notation. This framework, in the form of an optimization problem, is related by
\begin{align}
\label{eq:origopt}
    \min\limits_{\bx_{T}} & \hspace{.1cm} f(\bar{\bx}_{C}, H(\bar{\bx}_{C}, \bx_{T}), \bx_{T})\\\nonumber
    \text{s.t.} &\hspace{.1cm} \Psi(\bx_{T} - \bar{\bx}_{T}) \leq B\\\nonumber
    & \hspace{.1cm} l_t \leq x_{t} \leq u_t \text{ for }t \in T
\end{align}
where $f:\mathbb{R}^{p} \rightarrow \mathbb{R}$ is the real-valued output of an induced classification function, $H:\mathbb{R}^{|C|+|T|} \rightarrow \mathbb{R}^{|I|}$ is a function that estimates the indirectly changeable feature values \footnote{Practically speaking, $H$ is collection of trained regressors which have been each been induced in the ``usual'' way.}, $\Psi(\cdot)$ is a cost function, given by
\begin{align}
\label{eq:costfunc}
\Psi(\bz) = \sum\limits_{t \in T}c_{t}^{+}(z_{t})_{+}+c_{t}^{-}(z_{t})_{-}
\end{align}
that imposes feature-specific costs on the extent of deviation (i.e., change) from the original instance features values (denoted $\bar{\bx}_{T}$); $B$ is a budget that controls the amount of cumulative change allowed, and where $l_t$ and $u_t$ are feasibility bounds. In \eqref{eq:costfunc}, the function $(\cdot)_{+}$ is used when $z_t > 0$ and the function $(\cdot)_{-}$ when $z_t < 0$. These functions ensure that $z_t$ is a positive value and allow the inverse classification designer to imposed different costs on treatment value increases ($z_t > 0$ costs are specified by $c^{+}_t$) and decreases ($z_t < 0$ costs are specified by $c^{-}_t$). Also note that henceforth, for notational convenience, we will assume that $T = {1,2,\dots,|T|}$.

To further elaborate on the components of the framework, consider our running student performance example, namely the indirectly changeable features $I$, which include features such as aspirations for higher education. Practically speaking, one cannot simply adjust one's aspirations: one can, however, change the activities they engage in, which may have a cascading effect on such aspirations. Hence, we introduce $H$ to model these cascading dependencies. $H$ can be thought of as some function, composed of sub-functions, that takes as input a vector composed of control and treatment features and provides estimates for each of the indirectly changeable features; the sub-function estimators need not be the same, nor are the sub-functions necessarily the same (e.g.,~one might be a kernel regressor, while another a linear regressor). For example, in our experiments, we use a neural network to learn a joint mapping of the indirectly changeable features.

Furthermore, $f$ is induced using training data (i.e.,~$\{(\bx^{(i)},y^{(i)})\}_{i=1}^{n}$) in conjunction with some loss function $\mathcal{L}(\cdot)$, and some train-test-validation procedure (i.e., cross-validation, as in our experiments), to obtain an optimal hypothesis, which can be written
\begin{align}
\label{eq:optclassif}
f^* = \argmin\limits_{f \in \mathcal{G}} \left \{\mathcal{L}_{\text{CV}} \left (f(\bx^{(i)}),y^{(i)} \right ): i=1,\dots,n \right \}
\end{align}
where $\mathcal{G}$ is a hypothesis space, and where $\mathcal{L}_{\text{CV}}(\cdot)$ encompasses computing the loss over the folds of the cross-validation procedure and then retraining the optimal model to produce $f^*$. In this work $\mathcal{G}$ is defined over a variety of neural network architectures. Therefore, $f^{*}$ is optimized via \textbf{backpropagation}.

By contrast, instances are then optimized (i.e., inverse classification is applied) via \textit{inverse} backpropagation. That is, instead of taking the gradient w.r.t.~the parameters to minimize a loss function, as in
\begin{align}
\label{eq:backpro}
    \frac{\partial \mathcal{L}\left (f(\bx^{(i)}),y^{(i)} \right )}{\partial \mathtt{\mathbf{p}}}
\end{align}
where $\mathtt{\mathbf{p}}$ is assumed to be some arbitrary data structure (i.e., a tensor) organizing the parameters that are being optimized, the gradient is taken w.r.t.~the instance and real-valued classification function itself, as in
\begin{align}
\label{eq:invbp}
\nabla f \triangleq \frac{\partial f(\bx_{C}, H(\bx_{C}, \bx_{T}), \bx_{T})}{\partial \bx_{T}}.
\end{align}
Here, for the sake of convenience later, we denote this gradient with $\nabla f \in \mathbb{R}^{|T|}$.

\begin{table}[htp]
\centering
\begin{tabular}{ll}
\toprule
\textbf{Notation} & \textbf{Description} \\ \midrule
\vspace{.1cm}
$\bx^{(i)}_{C} \in \mathbb{R}^{|U|}$ & Control feature vector of instance $i$.\\
\vspace{.1cm}
$\bx^{(i)}_{I} \in \mathbb{R}^{|I|}$ & Instance $i$'s indirectly changeable feature \\
& vector.\\
$\bx^{(i)}_{T} \in \mathbb{R}^{|D|}$ & Treatment feature vector of instance $i$.\\
\vspace{.1cm}
$y^{(i)} \in \{0,1\}$ & Event label of instance $i$.\\\midrule
$f(\cdot)$ & Classification and objective function.\\
$\nabla f$ & Derivative of $f(\cdot)$ w.r.t.~$\mathbf{x}_T$.\\
$H(\cdot)$ & Indirectly changeable feature estimator.\\
$\Psi(\cdot)$ & Cost function.\\
$B$ & Budget.\\
$l_t \text{ } (u_t)$ for $t \in T$ & Lower (upper) bounds.\\
$\mathcal{L}_{\text{CV}}(\cdot)$ & Cross-validation performed over\\
& some loss function $\mathcal{L}$.\\\midrule
$\mathtt{\mathbf{g}}(\mathbf{x}_{C})$ & Gaussian process function.\\
$\mathbf{\Phi}(\mathbf{x}_T,\mathtt{\mathbf{g}}(\mathbf{x}_C))$ & Density estimate from the\\
& reconstructed predictive distribution (APS).\\\bottomrule
\end{tabular}
\caption{Notation used throughout this work.\label{tab:notation}}
\end{table}

\begin{figure}[htp]
    \centering
    \includegraphics[scale=0.50]{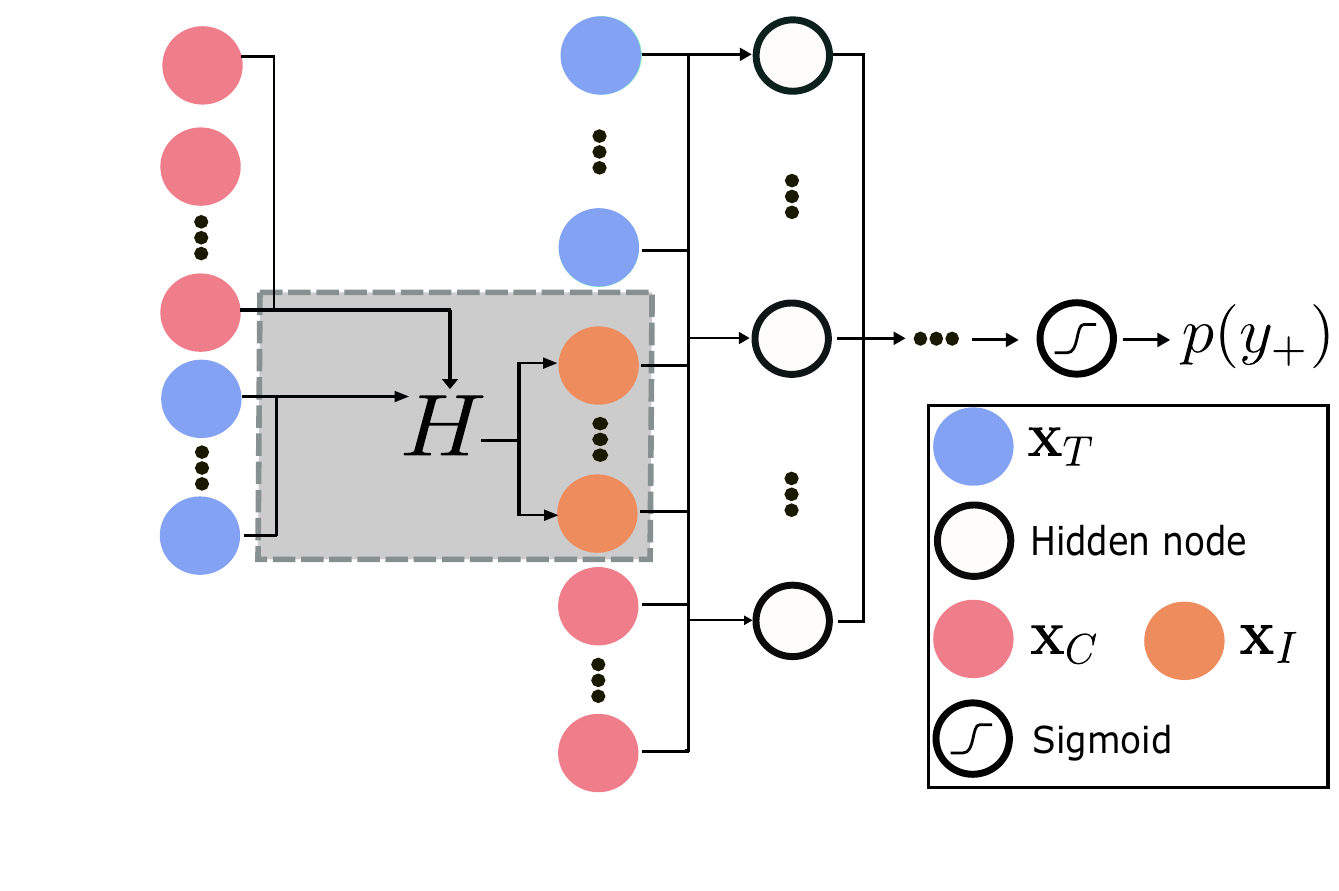}
    \caption{Non-causal neural network architecture, based on $f$.}
    \label{fig:nocause-arch}
\end{figure}

For the sake of clarity, the full inverse classification process is disclosed in Algorithm \ref{algo:icproc} and a graphic, showing the architecture of $f$ and use of $H$, is presented in Figure \ref{fig:nocause-arch}. We also provide some of the common notation used throughout this work in Table \ref{tab:notation}.

\begin{algorithm}[htp]
\begin{algorithmic}[1]
\caption{The Inverse Classification Process \label{algo:icproc}}
\REQUIRE Test instance $\bar{\bx}$, $\{(\bx^{(i)},y^{(i)})\}_{i=1}^{n}$, $c^{+}_{t}$, $c^{-}_{t}$, $l_t$, $u_t$ for $t \in T$.
\STATE Obtain $f^{*}$ and $H^{*}$ using Sub-
procedure \ref{subpro:learn} (this is done only once, not for every test instance).
\STATE Solve \eqref{eq:origopt} using \eqref{eq:invbp}.
\ENSURE Optimized instance $\bx^{*}$.
\end{algorithmic}
\end{algorithm}

\setcounter{algorithm}{0}
\begin{algorithm}
\floatname{algorithm}{Sub-procedure}
\begin{algorithmic}[1]
\caption{Induce $f^*$ and $H$ \label{subpro:learn}}
\REQUIRE $\{(\bx^{(i)},y^{(i)})\}_{i=1}^{n}$
\STATE Induce $f^{*}$ according to \eqref{eq:optclassif} using \eqref{eq:backpro}.
\STATE Induce indirect feature estimator $H^{*}$.
\ENSURE $f^{*}$ and $H^{*}$
\end{algorithmic}
\end{algorithm}

\subsection{Approximate Propensity Score-weighting}

The trouble with eliciting an optimal treatment policy (i.e.,~recommendations) from $f^*$ using \eqref{eq:origopt} is with the training data used to induce $f^{*}$. More specifically, because the training data is observational, \textit{selection bias} may obscure $f^{*}$ from learning the true treatment effects. Past works on eliciting causal effects from observational data have developed tools to address this issue by making certain assumptions and \textit{correcting} for selection bias. These works, however, have done so from a stand point of primarily single binary and single continuous treatments. In this work, however, we have multiple, continuously-valued treatments. As a result, we we must innovate with respect to these assumptions and selection bias corrections, found in past literature, to arrive at a suitable solution to the multiple, continuously valued nature of the treatments and causal inverse classification problem.

We begin by first disclosing the assumptions we are making. The first of these, Assumption \ref{assum:confound}, is referred to as the \textit{no unmeasured confounders} assumption.
\begin{assumption}
\label{assum:confound}
\textup{(No unmeasured confounders)} Assume that each of the treatments $x_t^{(i)}$ is conditionally independent of the outcome $y^{(i)}$ w.r.t.~the controls $\bx^{(i)}_{C}$.
\begin{align*}
    y^{(i)} \perp x_t^{(i)} | \bx^{(i)}_{C}: \text{ }t \in T
\end{align*}
\end{assumption}
Note that we have extended the well known single treatment formulation found in \cite{Pearl2009,Hirano2004}, as well as others that we relate in Section 4. This assumption ensures that the \textit{treatment assignment} (more on this shortly) is independent of the outcomes. This is a necessary assumption for estimating \textit{treatment propensity}, which will be discussed shortly.

Assumption \ref{assum:nonz} is referred to as the \textit{non-zero treatment probability} assumption.
\begin{assumption}
\label{assum:nonz}
\textup{(Non-zero treatment probability)} Assume that the probability of observing any particular treatment value $x_t^{(i)}$ for a particular treatment indexed by $t$ is non-zero.
\begin{align*}
\{0 < P(x_t^{(i)}|\bx_{C}^{(i)}): t \in T\}
\end{align*}
\end{assumption}
Coupled together, Assumptions \ref{assum:confound} and \ref{assum:nonz} form the \textit{strong ignorabilitiy conditions} of  Rosenbaum and Rubin (1983a) \cite{Rosenbaum1983}, adapted to our multiple, continuously valued treatment formulation\footnote{The Rosenbaum and Rubin assumptions were formulated based on a single, binary treatment, but can be generalized to single, continuously valued treatments \cite{Pearl2009}.}.

These conditions were originally formulated with only a single treatment in mind. Therefore, we find the need for an additional assumption, \textit{independent treatments, conditioned on controls (ITCC)}, presented in Assumption \ref{assum:indT}.
\begin{assumption}
\label{assum:indT}
\textup{(Independent treatments, conditioned on controls (ITCC))} Assume that the treatment value $x_t^{(i)}$ for treatment $t$ is independent of treatment value $x_k^{(i)}$ for treatment $k$ w.r.t.~the controls.
\begin{align*}
    x_t^{(i)} \perp x_k^{(i)}|\bx^{(i)}_{C}: \text{ }t,k \in T, t\neq k
\end{align*}
\end{assumption}
The need for Assumption \ref{assum:indT} stems from the need to estimate the assignment mechanism of each treatment $t \in T$ independently of treatment $k \in T$ where $t \neq k$. Collectively, we refer to this updated set of assumptions as the \textbf{multiple treatment - strong ignorability assumptions}.

\subsubsection{Assignment Mechanism: Gaussian Processes}

In disclosing the multiple treatment - strong ignorability assumptions we made several allusions to the \textit{assignment mechanism}. The \textbf{assignment mechanism} refers to the underlying process governing treatment assignment (i.e., how/why the values of a particular instance are observed as they are). This process can best be illustrated through the juxtaposition of a randomized controlled trial (RCT) and the nature of observational data. In the former setting (RCT), the so called treatment assignment is completely random w.r.t.~to the population, characterized by so-called control covariates (e.g.,~age), comprising the study (hence \textit{randomized} trial). Therefore,
\begin{align*}
    \{P(x^{(i)}_{t}|\bx^{(i)}_{C}) = P(x^{(l)}_{t}|\bx^{(l)}_{C}): \text{ for }i \neq l\}.
\end{align*}
However, in observational data
\begin{align*}
    \{P(x^{(i)}_{t}|\bx^{(i)}_{C}) \neq P(x^{(l)}_{t}|\bx^{(l)}_{C}): \text{ for } i \neq l\}.
\end{align*}
In other words, assignment is not random: this is the aforementioned problem of selection bias. Therefore, the task in an observational data setting is to uncover the assignment mechanism such that treatment propensities (\textit{propensity scores}), the probability of an instance receiving the particular treatment (amount) they currently are, can be obtained. Subsequently, the propensity score is used as a normalizing weight to correct for assignment mechanism (selection) bias when inducing classifiers and thus eliciting classifiers that are causal in nature (we relate these works in Section 4).

These past propensity-based correction methods are, however, unsuitable for our purposes because of the multiple, continuously valued treatments inherent to our problem formulation. We use these past works as motivation and adapt our own methods to the state of the art to account for the multiple, continuously valued nature of our treatments.

In this work, we make the following assumption as to the treatment assignment mechanism
\begin{assumption}
\textup{(Gaussian Process-based Treatment Assignment)} We assume that the assignment mechanism $\forall t \in T$ is governed by a Gaussian process among the control features. This is formally expressed as
\begin{align}
\label{eq:gauss}
    \mathcal{\mathbf{g}}(\bx_{C}) = \left[ \text{GP} \left( \mu_{t}(\bx_{C}),\kappa_{t}(\bx_{C},\bx_{C}^{(i)}) \right) \right]_{t \in T}\text{ : }i=1,\dots,n
\end{align}
where $\kappa_t(\cdot)$ is a specified kernel function, for the treatment indexed by $t$, used to compute the covariance matrix, and $\mu_t(\cdot)$ is a specified mean function. The Gaussian process function $\text{GP}$ uses these elements, along with optimized hyperparameters, to make predictions for test instances.
\end{assumption}
Our assumption about the treatment assignment is one of both flexibility and convenience: the Gaussian process kernel function is incredibly flexible and capable of modeling nonlinearities that may exist among the control features \cite{Rasmussen2004}. Therefore, we can forgo assuming independence among the control features themselves. Gaussian processes are further convenient for two primary reasons:
\begin{itemize}
    \item The estimate has immediate maximum likelihood characteristics.
    \item The predictive distribution can be re-constructed about the estimated point and is Gaussian (a differentiable and $L$-Lipshitz continuous gradient function).
\end{itemize}

While the point of convenience is not fully realized until the next subsection (the differentiability and continuous gradient of the Gaussian is optimizationally convenient), we relate the reconstructed predictive distribution, used to elicit the density estimate, by
\begin{align}
\label{eq:preddist}
    \Phi(\bx_{T},\mathtt{\mathbf{g}}(\bx_{C})) = \left[\frac{1}{\sqrt{2 \pi}\mathtt{\mathbf{g}}^{\sigma}_{t}} \cdot \text{exp}\left( \frac{- \left(x_{t}-\mathtt{\mathbf{g}}^{x}_{t} \right)^2}{2 \mathtt{\mathbf{g}}^{\sigma^2}_{t}} \right) \right]_{t \in T}
\end{align}
where $\mathtt{\mathbf{g}}^{x}_{t}$ is the estimate elicited from the Gaussian process function, and $\mathtt{\mathbf{g}}^{\sigma}_{t}$ is the standard deviation elicited from the covariance matrix. Practically speaking, $\Phi(\cdot)$ produces a density estimate of the instance's current treatment value $x_t$ relative to the Gaussian process predicted value $\mathtt{\mathbf{g}}^{x}_t$. This density estimate is the \textbf{approximate propensity score} (APS) and will be used to weight each instance's respective treatment vector to produce a weighted treatment, much like inverse propensity score weighting, which can be expressed formally by
\begin{align}
    \mathbf{x}_{T}^{\prime} \triangleq \Phi(\bx_{T},\mathtt{\mathbf{g}}(\bx_{C})) \odot \bx_{T}
\end{align}
where $\odot$ denotes element-wise multiplication between each of the $t \in T$ treatment values and corresponding APS. The intuition is that instances receiving a treatment close to the unbiased estimate of the treatment we would expect them to receive will have inflated values, while instances receiving a treatment far away from the unbiased estimate will have their weighted treatment value pushed toward 0. In the next subsection we discuss how we can update the inverse classification process to trade-off optimizing for causality and probabilistic improvement by taking into account the gradient information of the APS.

\begin{figure}[h]
    \centering
    \includegraphics[scale=0.50]{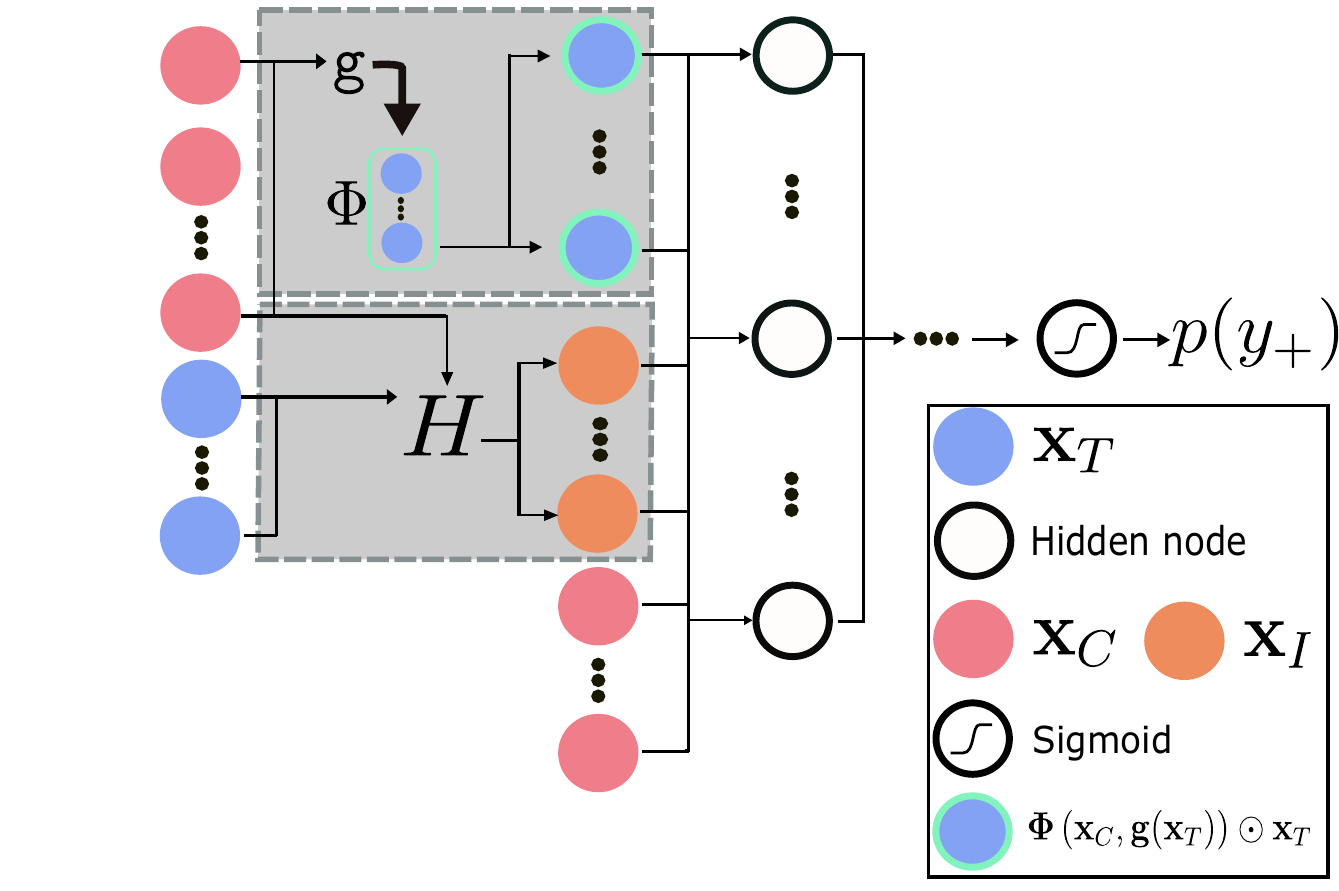}
    \caption{Causal neural network architecture, based on $f$.}
    \label{fig:cause-arch}
\end{figure}

For convenience we present an updated neural network architecture in Figure \ref{fig:cause-arch}.

\subsection{Individual-specific Treatment Policy Optimization}

With the APS-weighting scheme disclosed, we return to discuss the original optimization we proposed and formalized in \eqref{eq:origopt}. Namely, we wish to incorporate the APS-weighting scheme into the optimization. Initially, we can update the formulation of \eqref{eq:origopt} to
\begin{align}
\label{eq:newobj}
    f^{\prime}(\bx) = f \left( \dots, \bx_{T}^{\prime} \right) = f \left(\dots, \textcolor{red}{\Phi \left( \bx_{T},\mathcal{\mathbf{g}} \right) \odot \bx_{T}} \right).
\end{align}
Subsequently, however, we express two different optimization updates. The first we express as
\begin{align}
\label{eq:newobjopt}
    \bx_{T}^{m+1} =   \Gamma \left(\bx_{T}^{m} -\eta \nabla f^{\prime} \right)
\end{align}
where $m$ is the current iteration of the gradient descent-based optimization process, $\Gamma(\cdot)$ is the projection operator that projects the update of \eqref{eq:newobjopt} to the feasible domain of \eqref{eq:origopt} (corresponding to projected gradient descent), $\eta$ is the step size and $\nabla f^{\prime}$ is the gradient of \eqref{eq:newobj} w.r.t.~ $x_T^{\prime}$. The issue with \eqref{eq:newobjopt} is that it is functionally a black box, as there is no gradient information specifying how updating $\bx_T$ effects $\Phi(\bx_{T},\mathtt{\mathbf{g}}(\bx_{C}))$. Hence, we specify a second optimization procedure
to include information about the APS (by applying the chain rule), thus allowing us to also optimize for propensity, thus alleviating the black box issue. The optimization updates can now be written
\begin{align}
\label{eq:fprimeopt}
    \bx_{T}^{m+1} = \Gamma \left( \bx_{T}^{m} - \eta \left[ \nabla f^{\prime} \textcolor{red}{\cdot (\Phi +\nabla \Phi \odot \bx_{T})}   \right] \right)
\end{align}
with the addition, differentiating \eqref{eq:newobjopt} from \eqref{eq:fprimeopt}, also highlighted in \textcolor{red}{red}, and for the sake of clarity we refer to $\Phi \left( \bx_{T},\mathcal{\mathbf{g}} \right)$ and the first-order derivative as $\Phi$ and $\nabla \Phi$, respectively.

However, there are still two potential issues with \eqref{eq:fprimeopt}. First, the APS is nonlinear, which may pose problems during the optimization. Second, it may not always be beneficial to completely optimize w.r.t.~the APS and we would, therefore, like to trade off optimizing for the APS and potential probability improvement (i.e., optimizing for the APS may sacrifice improvement in the classification score elicited from $f$). To such an end, we formulate a new objective function, $g$, that imposes a regularizer limiting the extent to which the APS is optimized.  This is related by
\begin{align}
\label{eq:g}
    g(\bx) = f \left( \dots, \textcolor{red}{\Phi \left( \bx_{T},\mathtt{\mathbf{g}} \right) \odot \bx_{T}} \right)  \textcolor{blue}{+ \lambda \left\lVert\frac{\bx_{T}-\mathtt{\mathbf{g}}^{x}}{2 \mathtt{\mathbf{g}}^{\sigma^2}} \right\rVert^2}
\end{align}
where $\lVert \cdot \rVert^2$ is the Euclidean norm squared and $\lambda$ is the regularizer; the updates are highlighted in \textcolor{blue}{blue}. The reason for imposing the Euclidean norm squared, instead of using the actual gradient of the APS Gaussian (i.e., reconstructed predictive distribution) is precisely because of the nonlinear nature of the Gaussian: by taking the log of the derivative of the Gaussian, as is typically done, we arrive at a single-side optimization and, by subsequently using the Euclidean norm squared, we can optimize regardless of which side the test point falls on. 

We update the optimization procedure to reflect \eqref{eq:g}, which is related by
\begin{align}
\label{eq:gopt}
    \bx_{T}^{m+1} =  \Gamma \left(\bx_{T}^{m} - \eta \left[ \nabla f^{\prime} \textcolor{red}{\cdot (\Phi +\nabla \Phi \odot \bx_{T})}  \textcolor{blue}{+ \lambda \frac{\bx_{T}-\mathtt{\mathbf{g}}^{x}}{\mathtt{\mathbf{g}}^{\sigma^2}}} \right] \right).
\end{align}

For convenience, we update Algorithm \ref{algo:icproc} to reflect the causal APS-weighting scheme we've disclosed in Algorithm \ref{algo:icprocg}.

\begin{algorithm}[htp]
\begin{algorithmic}[1]
\caption{The Causal Inverse Classification Process \label{algo:icprocg}}
\REQUIRE Test instance $\bar{\bx}$, $\{(\bx^{(i)},y^{(i)})\}_{i=1}^{n}$, $\lambda$, $c^{+}_{t}$, $c^{-}_{t}$, $l_t$, $u_t$,$\mu_t$, $\kappa_t$ for $t \in T$.
\STATE Obtain $\bx_{T}^{(i)\prime}$ by weighting $\bx_{T}$ according to \eqref{eq:preddist}, using \eqref{eq:gauss}.
\STATE Obtain $f^{*}$ and $H^{*}$ using Sub-
procedure \ref{subpro:learn} (this is done only once, not for every test instance).
\STATE Solve \eqref{eq:g} using \eqref{eq:gopt}\footnotemark.
\ENSURE Optimized instance $\bx^{*}$.
\end{algorithmic}
\end{algorithm}
\footnote[4]{We explore using $f^{\prime}$ without APS gradient, $f^{\prime}$ with APS gradient and $g$ in our experiments in Section 4.}

\subsection{Inverse Classification Potential Outcomes Framework}

Up to this point we have discussed inverse classification, our proposal to update the framework, and a process to elicit causal recommendations -- i.e.,~treatment policies composed of multiple, continuously valued treatments. In this subsection, we update the familiar \textit{potential outcomes framework} (POF) to account for:
\begin{enumerate}
    \item The multiple, continuously valued treatments.
    \item The inverse classification formulation.
\end{enumerate}
By updating this framework, the bridge between causal inference and inverse classification should become clearer.

In a single, binary treatment setting, the individual treatment effect (ITE) can be written
\begin{align}
    \label{eq:binite}
    ITE = Y_i(x_t=1) - Y_i(x_t=0)
\end{align}
where $Y_i(\cdot)$ is a function that determines the outcome provided the treatment input, and where $x_t \in \{0,1\}$ is a binary, scalar input; the difference in the two provided quantities -- the outcome when the treatment is and isn't taken -- is referred to as the \textit{individual treatment effect}.

However, one only ever observes either $Y_i(x_t=1)$ or $Y_i(x_t=1)$, but never both. The process of determining this unknown quantity is referred to as \textit{counterfactual inference}; the unknown outcome is referred to as the counterfactual.
To reflect this, we update \eqref{eq:binite} to yield the \textit{potential outcomes framework}, given by
\begin{align}
\label{eq:ite}
    ITE = \left\{
    \begin{array}{ll}
    Y(x_{t}=1) - \hat{Y}(x_t=0) & \text{If }t_1\text{ is observed.}\\
    \hat{Y}(x_t=1) - Y(x_t=0) & \text{If }t_0\text{ is observed.}\\
    \end{array}
    \right.
\end{align}

For multiple, continuously-valued treatments, we propose the following \textbf{multiple, continuously valued potential outcomes framework} ($m$PO), given by
\begin{align}
\label{eq:mPO}
    ITE = Y(\bar{\bx}_{T}) - \hat{Y}(\bx_{T}^{*})
\end{align}
where we are taking the known outcome with the observational data (i.e.,~$Y(\bar{\bx}_{T})$) and obtaining some optimized policy for which an estimate is provided (i.e.,$\hat{Y}(\bx_{T}^{*})$).

While PO and $\mathtt{m}$PO are helpful for causal inference -- that is, to infer \textit{what} causally effects the outcome of interest, relative to the known outcome -- in an inverse classification setting, \textbf{neither} outcome is observed: we do not know what will happen if the patient continues their current course of action or if they adopt the suggested treatment policy. Therefore, we update \eqref{eq:mPO} to the \textbf{inverse classification potential outcomes framework} (ICPOF), using what we term as the \textbf{individual future estimated effects} ($i$FEE), given by
\begin{align}
    i\text{FEE} = f^{\prime}(\bar{\bx}_{T}) - f^{\prime}(\bx_{T}^*)
\end{align}
where we substitute $f^{\prime}(\cdot)$\footnote{Note that we substitute $f(\cdot)$ for $f^{\prime}(\cdot)$ when doing non-causal comparison experiments in Section 4.} for both $Y(\cdot)$ and $\hat{Y}(\cdot)$ to reflect the fact that estimation must take place for both the instance's current treatment values and for the optimized treatment values. The intuition is simple: we are trying to optimize for some unknown future outcome, therefore requiring estimation as to outcome probability.

In our experiments, discussed in the next section, we assess success using average $i$FEE among the test set.

\section{Experiments and Results}

In this section we present our experiment methodology, as well as the datasets, parameters and results of our experiments. 

\subsection{Experiment Methodology and Datasets}

As we mention at the end of Section 3, traditional means of assessing treatment effects at the individual level (i.e., ITE) are unsuitable for our purposes, as we are interested in estimating future outcomes.  Therefore, we develop $i$FEE -- individual future estimated effects -- for assessing the success of optimized individual treatment policies.

In our experiments, we explore the use of four different objective functions: non-causal $f$, which shows a result obtained from a non-causal inverse classification process, $f^{\prime}$ without the APS-incorporated gradient (abbreviated $f^{\prime}$-no opt in our experiments), which uses \eqref{eq:newobj} and \eqref{eq:newobjopt} as the objective function and optimization update, respectively, $f^{\prime}$ with the APS-incoporated gradient (abbreviated  $f^{\prime}$-opt in our experiments), which uses \eqref{eq:newobj} and \eqref{eq:fprimeopt}, and $g$, which uses \eqref{eq:g} and \eqref{eq:gopt}. To evaluate the success of each function we report the average $i$FEE among all testing data for each model across a variety of budgets $B$. Note that average $i$FEE is different than \textit{average treatment effect} (ATE) of other causal works, as ATE computes the effect of a single policy across the study population, while we are evaluating individual-specific policies; hence, average $i$FEE.

We, however, do not want to use the same model to evaluate the success of each optimized policy as was used to perform the optimization. Therefore, borrowing from the validation procedure disclosed in \cite{Lash2017b}, we induce two models: an optimization model and a biased validation model. To do this, we split our dataset in half, using the first to train an optimization model and the second to train a validation model. We then use the optimization model to perform inverse classification on each of the instances used to induce the validation model. Subsequently, we obtain the $i$FEE using the validation model. Note that the bias implicit in this validation procedure is intentional: the elicited effects that we report will be biased towards each $\bar{\bx}$, as those instances were used to induce the validation model. The reason for this intentional bias is to err on the side of caution in reporting our results, and should make for a more convincing argument as to the success of our outlined procedure. For the sake of clarity, we report this evaluative process in Algorithm \ref{algo:exps}.

\begin{algorithm}[htp]
\begin{algorithmic}[1]
\caption{Experiment Process \label{algo:exps}}
\STATE Split $\{(\bx^{(i)},y^{(i)})\}_{i=1}^{n}$ in half: $\bX^{\text{opt}}$ and $\bX^{\text{val}}$.
\STATE Use Sub-procedure \ref{subpro:learn} with $\bX^{\text{opt}}$ to induce $f^{*}_{\text{opt}}$ and $H^{*}_{\text{opt}}$ (also learn Gaussian processes among the $T$ features).
\STATE Use Sub-procedure \ref{subpro:learn} with $\bX^{\text{val}}$ to induce $f^{*}_{\text{val}}$ and $H^{*}_{\text{val}}$.
\STATE For each instance $\bar{\bx} \in \bX^{\text{val}}$, perform inverse classification using $f^{*}_{\text{opt}}$ and $H^{*}_{\text{opt}}$ with the corresponding objective function ($f$, $g$, $f^{\prime}$) to obtain $\bx^{*}$.
\STATE Evaluate $\bar{\bx}$ and $\bx^{*}$ using $f^{*}_{\text{val}}$ and $H^{*}_{\text{val}}$ and report the $i$FEE.
\end{algorithmic}
\end{algorithm}

\begin{figure*}[htp]
    \centering
    \begin{subfigure}[]{.49\linewidth}
        \centering
        \captionsetup{justification=centering}
        \includegraphics[scale=.45]{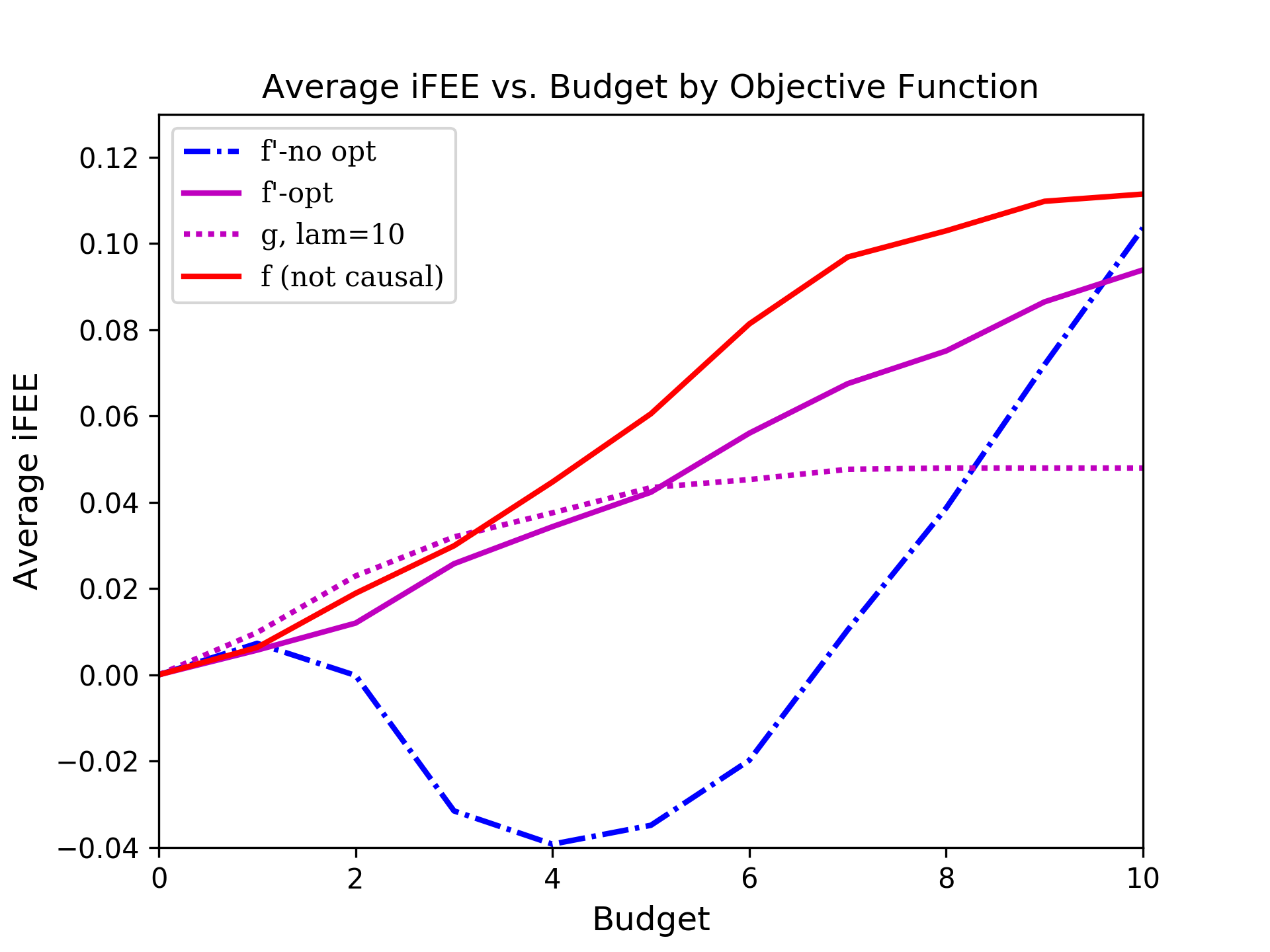}
        \caption{Avg.~Probability improvement by obj.~function.\label{fig:imp-bench}}
    \end{subfigure}
    \begin{subfigure}[]{.49\linewidth}
        \centering
        \captionsetup{justification=centering}
        \includegraphics[scale=.45]{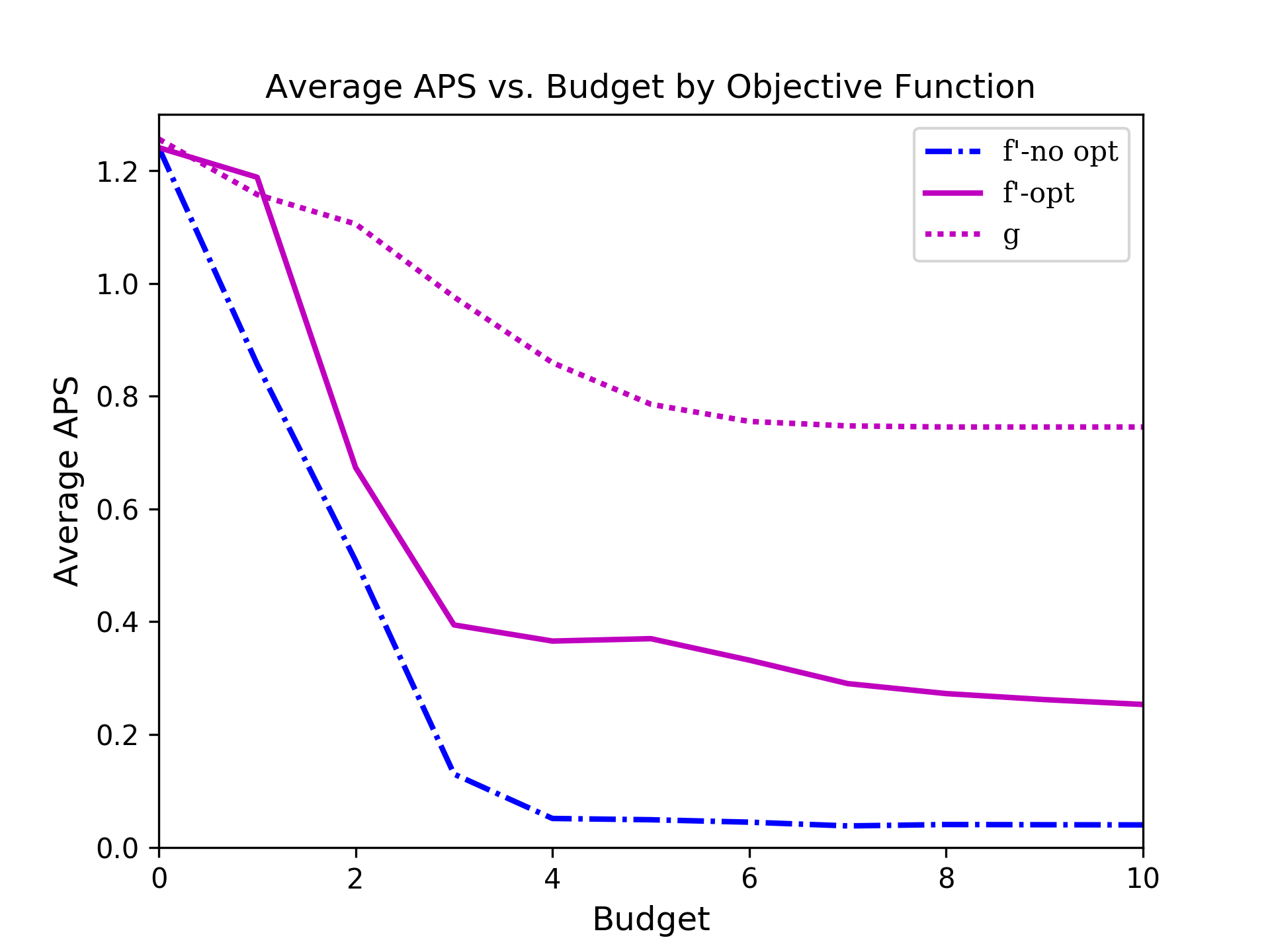}
        \caption{Avg.~APS Among Treatment Policies.\label{fig:impaps-bench}}
    \end{subfigure}
    \caption{Average $i$FEE and APS.\label{fig:improve}}
\end{figure*}

\subsubsection{Datasets}

In our experiments, we validate our methods using student performance, which is a freely available dataset from the UCI Machine Learning Repository. Our processed dataset student performance datasets consists of $649$ instances and $43$ features. Processing consisted of binarization of categorical features and normalization of the data. The $y^{+}$ class of interest is defined to be a grade of C or worse for student performance. \textbf{We have made our code, processed dataset, and experiments freely available to the public at {\sffamily \textcolor{dgreen}{https://github.com/michael-lash/Prophit}}}. Our treatment features consist of daily alcohol consumption, weekend alcohol consumption, study time, amount of time spent with friends, and paid tutoring. The full description of the data, $C$ and $I$ feature sets, and cost vectors can be found at {\sffamily \textcolor{dgreen}{https://github.com/michael-lash/Prophit}}. 

We wish to point out that, due to the lack of suitable, freely available datasets (e.g., lacking the $C,I,T$ feature sets, necessary continuous treatments, etc.) we have restricted ourselves to using a single dataset, which we experimentally explore thoroughly. We hope this paper, and our publicly available code, will encourage researchers to use our developed methodology to explore and subsequently make available new datasets that can be used to validate this and other methods building upon our work.

\subsection{Results}

In this subsection, we report the results of our experiments. First, we present the average $i$FEE by objective function ($f$, $f^{\prime}$-no opt, $f^{\prime}$-opt, and $g$) and budget value, along with average APS among individual treatment policies\footnote{The average is first computed instance-wise using only those APS' that comprise the individual's treatment policy. An average is then take among those instance-based averages that fall within three standard deviations of the overall mean.}. Second, we present results showing calibration of the  $\lambda$ values in $g$ along with corresponding average APS results. Finally, we show the most common treatment recommendations by objective function ($f$, $f^{\prime}$-no opt, $f^{\prime}$-opt, and $g$).

\subsubsection{Average $i$FEE by Objective Function}

We present the average $i$FEE results by objective function and by budget level in Figure \ref{fig:improve}.


In examining Figure \ref{fig:imp-bench}, we first point out that all objective functions were able to produce an improvement in average $i$FEE. This observation has several accompanying caveats, however. First, the non-causal method performed the best at higher budgetary levels. This result is, of course, non-causal in nature and is unreliable. Second, the $g$ objective function performs the best for the first three budgetary constraints ($B=1,2,3$), which is very encouraging. The result begins to asymptote beginning at a budget $B=4$, which also corresponds to the beginning of asymptotical average APS values, shown in Figure \ref{fig:impaps-bench}. This shows that $g$ is able to maintain an ample level of propensity, while still being able to provide $i$FEE-improving treatment policies. Unsurprisingly, $f^{\prime}$-no opt struggled to produce probability improvements, even producing policies that led to worse $i$FEE, initially. The average APS of $f^{\prime}$-no opt quickly falls to near-zero values, as shown in Figure \ref{fig:impaps-bench}, indicating untrustworthy results (in terms of causality). Finally, $f^{\prime}$-opt represents a moderate result. At lower budget levels, the result is slightly worse than $g$, while at larger $B$ levels is better. As we an see in Figure \ref{fig:impaps-bench}, the average APS of $f^{\prime}$-opt falls significantly at larger budgetary levels and can only be trusted moderately.

\subsubsection{$\lambda$ calibration}

\begin{figure*}[htp]
    \centering
    \begin{subfigure}[]{.19\linewidth}
        \centering
        \captionsetup{justification=centering}
        \includegraphics[scale=.22]{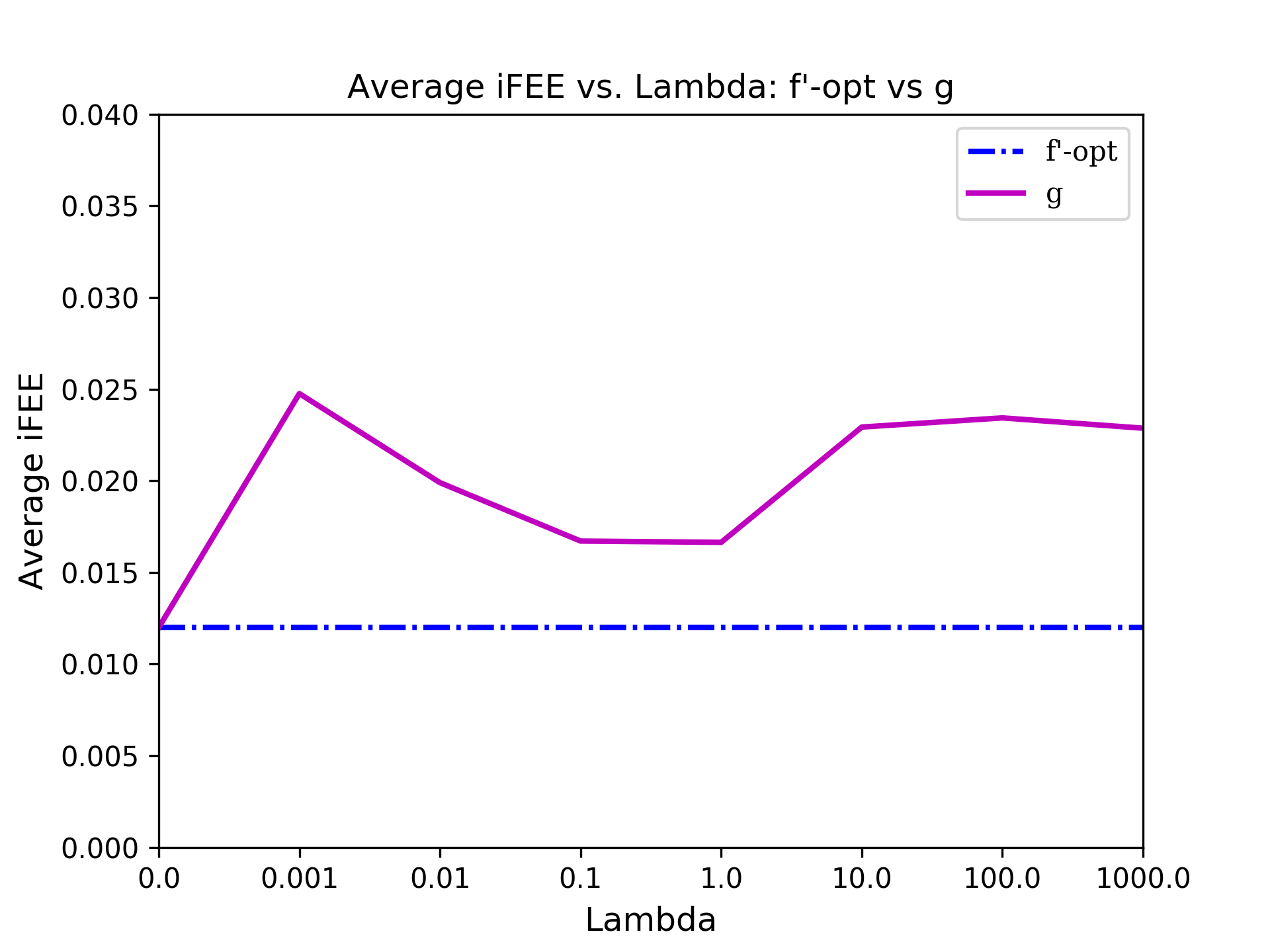}
        \caption{$B=2$.\label{fig:lam-benchb2}}
    \end{subfigure}
    \begin{subfigure}[]{.19\linewidth}
        \centering
        \captionsetup{justification=centering}
        \includegraphics[scale=.22]{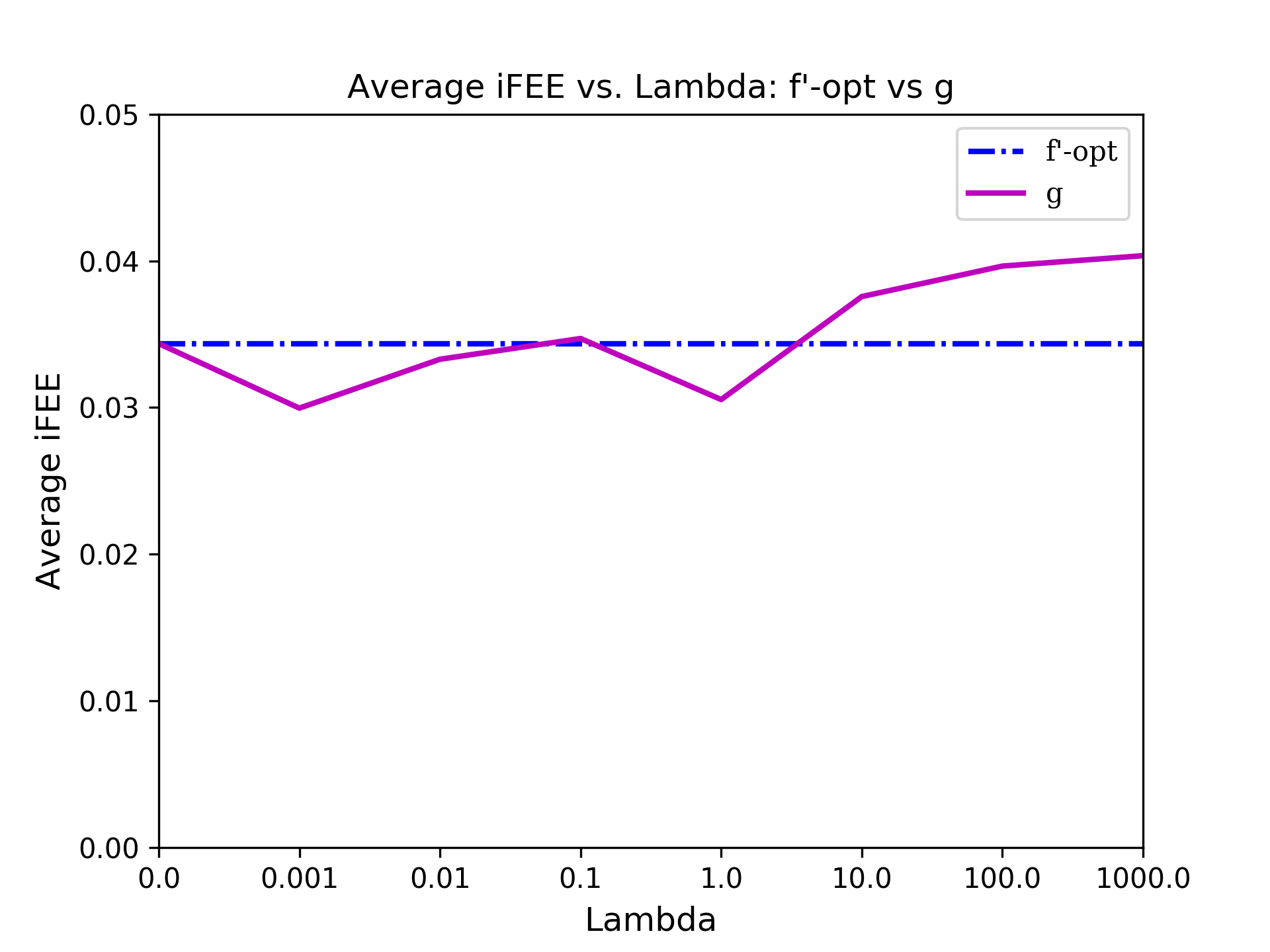}
        \caption{$B=4$.\label{fig:lam-benchb4}}
    \end{subfigure}
    \begin{subfigure}[]{.19\linewidth}
        \centering
        \captionsetup{justification=centering}
        \includegraphics[scale=.22]{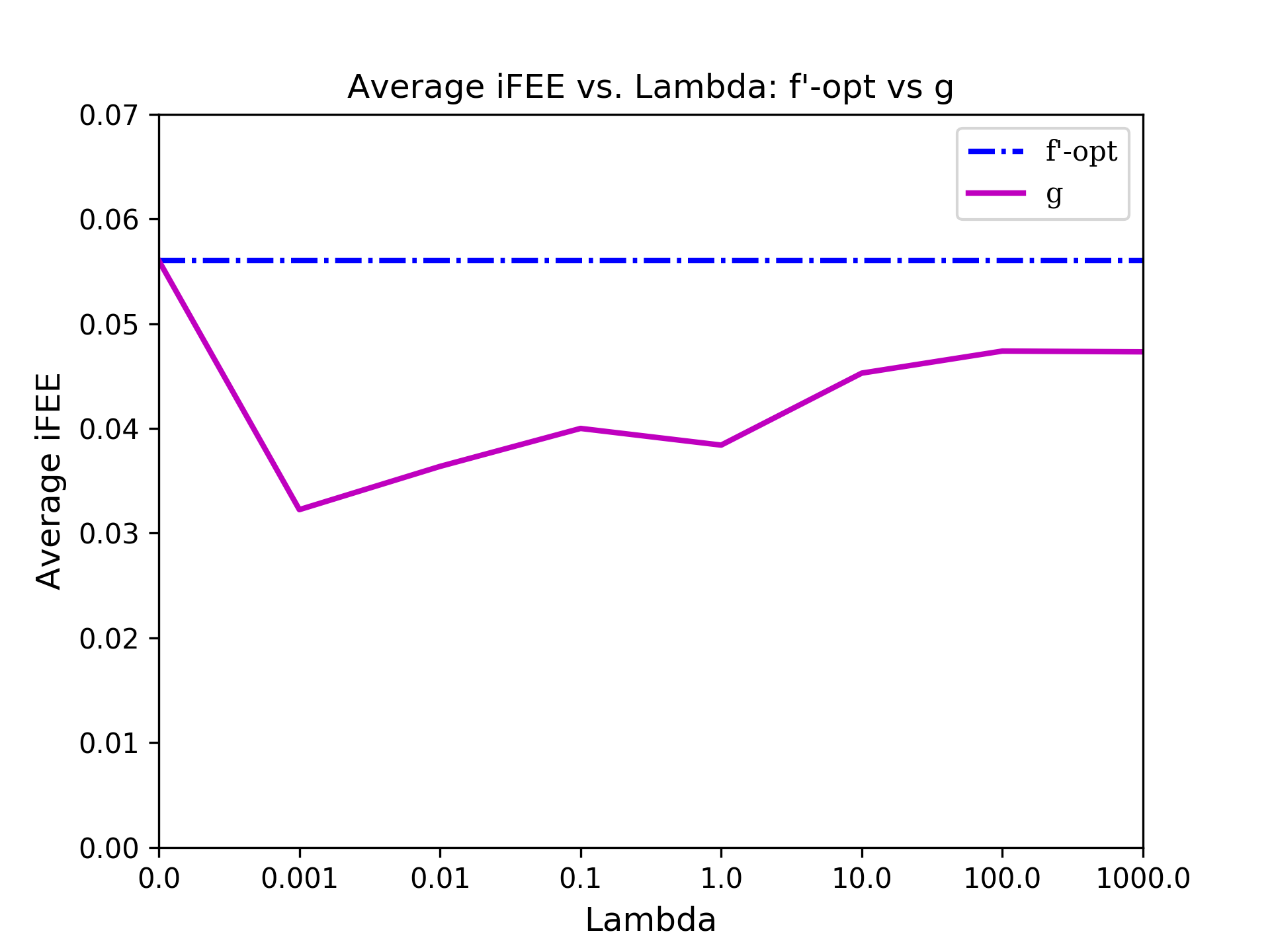}
        \caption{$B=6$.\label{fig:lam-benchb6}}
    \end{subfigure}
    \begin{subfigure}[]{.19\linewidth}
        \centering
        \captionsetup{justification=centering}
        \includegraphics[scale=.22]{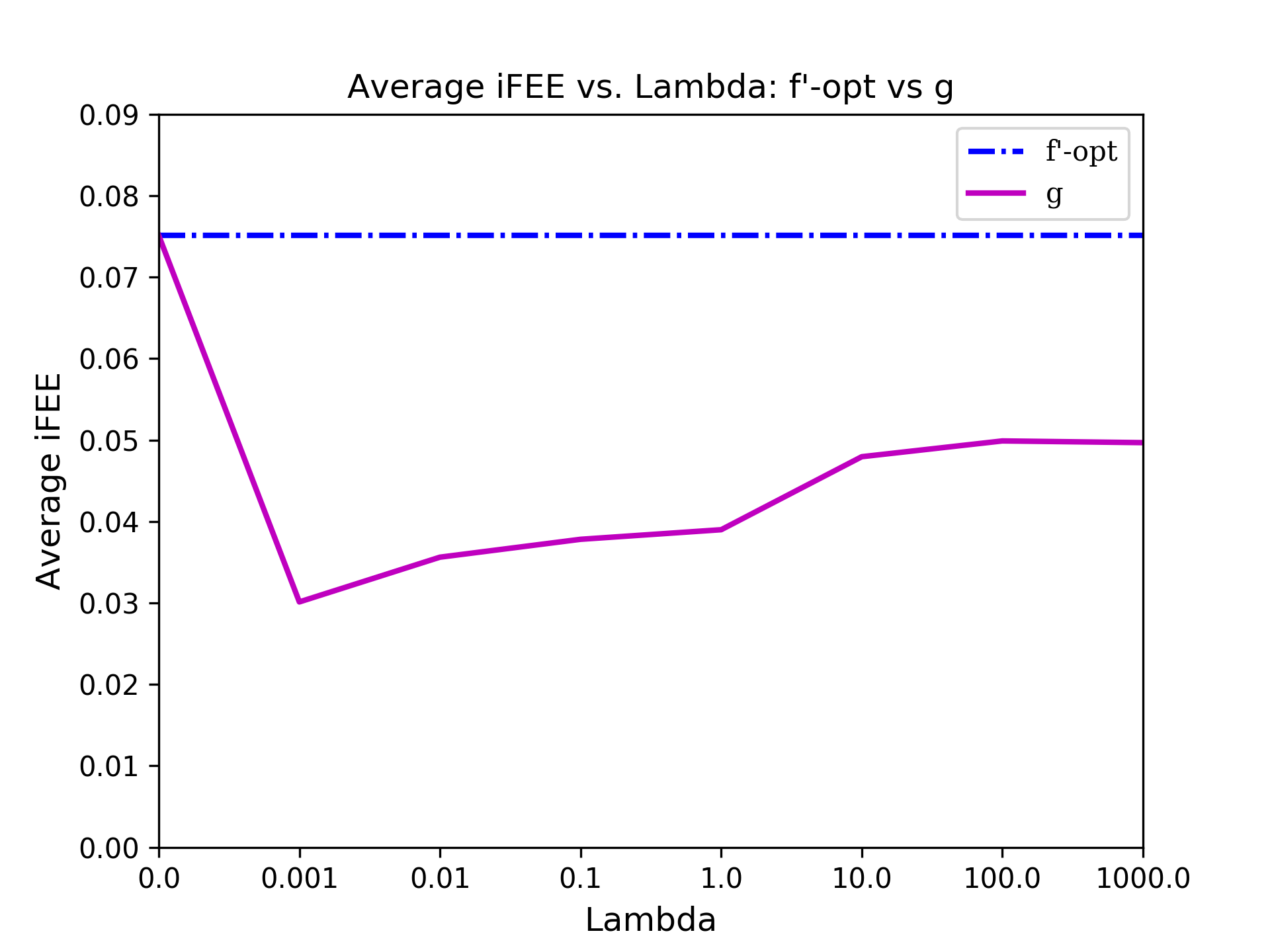}
        \caption{$B=8$.\label{fig:lam-benchb8}}
    \end{subfigure}
    \begin{subfigure}[]{.19\linewidth}
        \centering
        \captionsetup{justification=centering}
        \includegraphics[scale=.22]{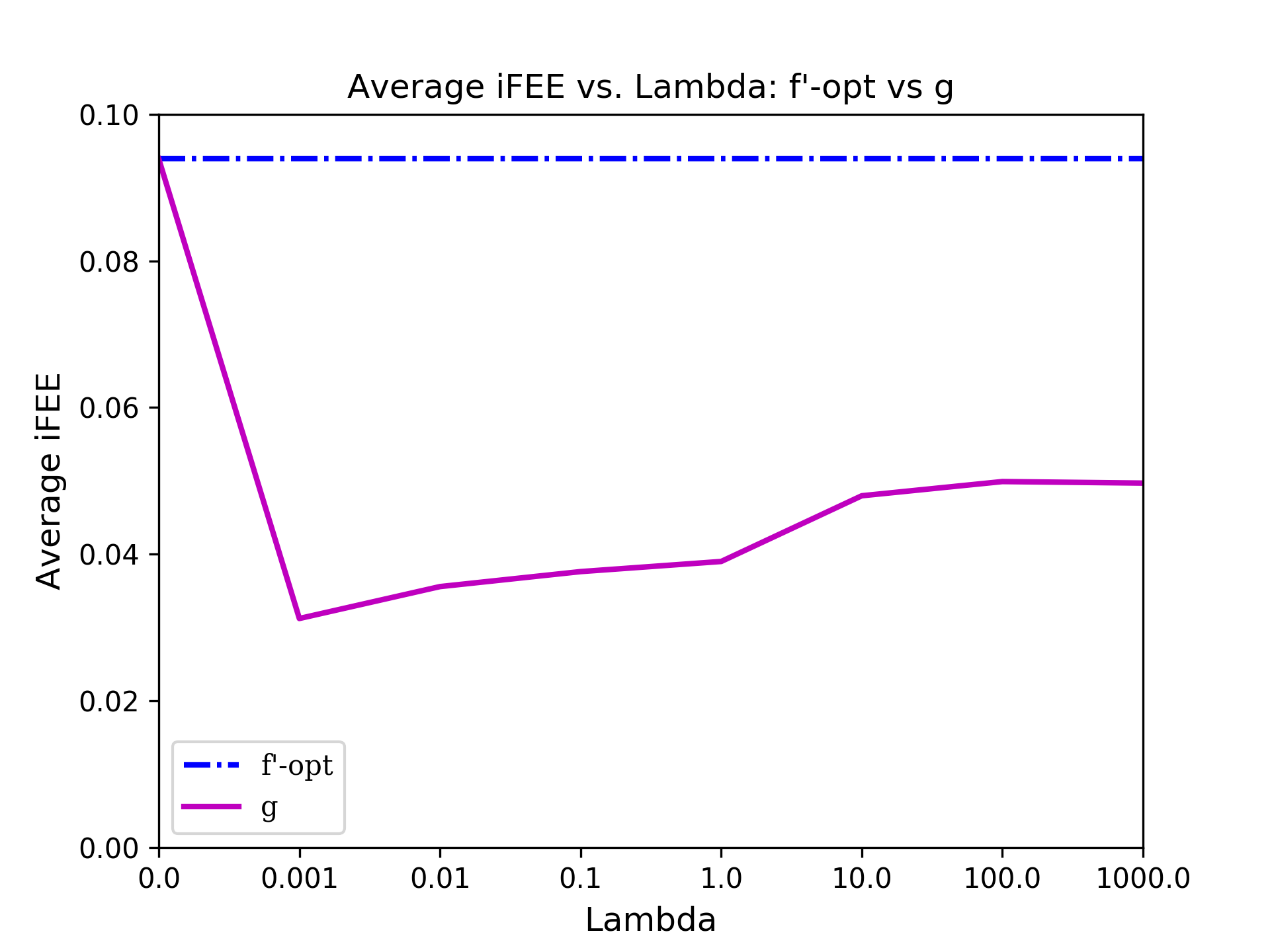}
        \caption{$B=10$.\label{fig:lam-benchb10}}
    \end{subfigure}\par
    \begin{subfigure}[]{.19\linewidth}
        \centering
        \captionsetup{justification=centering}
        \includegraphics[scale=.22]{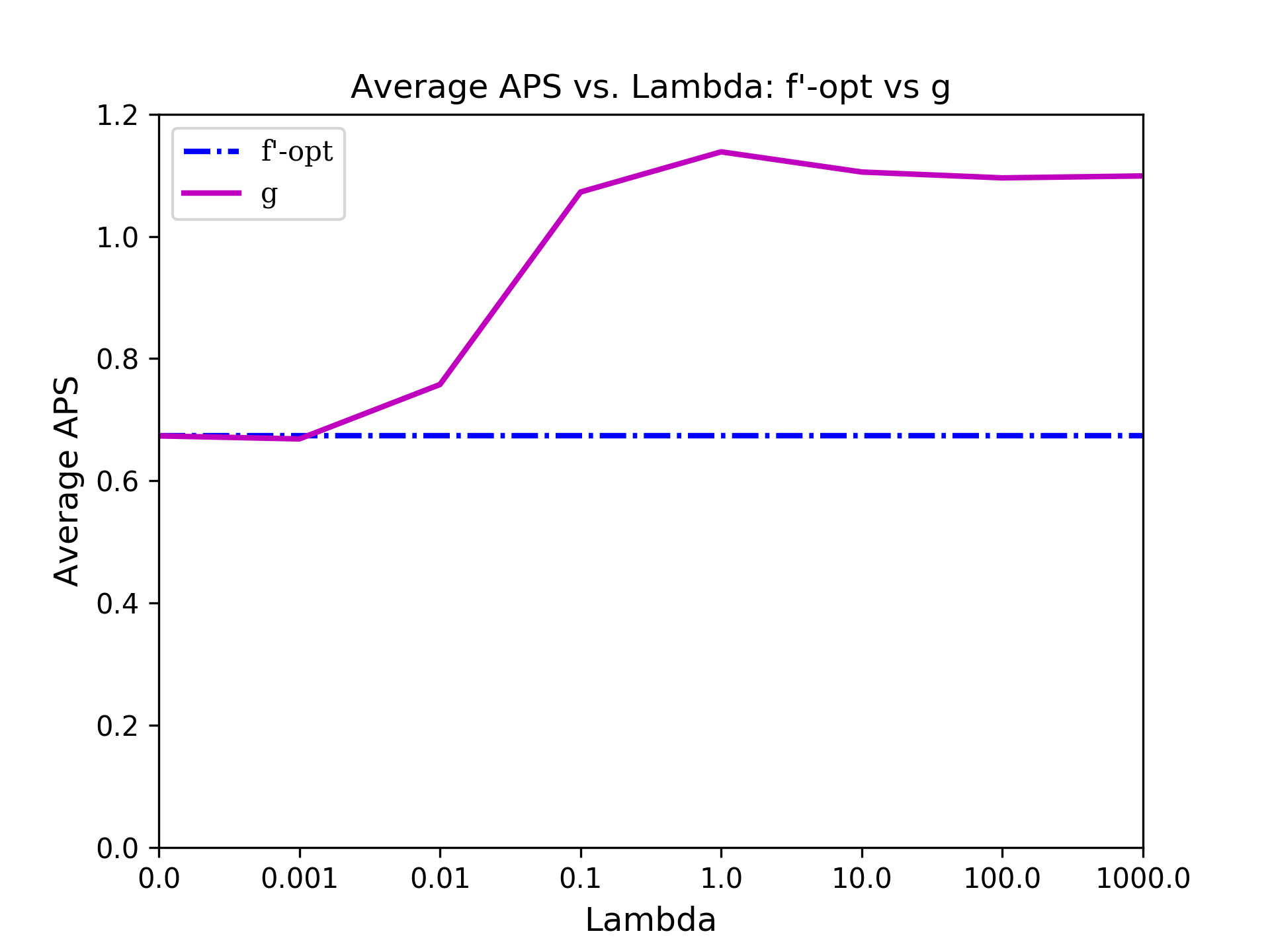}
        \caption{$B=2$.\label{fig:aps-benchb2}}
    \end{subfigure}
    \begin{subfigure}[]{.19\linewidth}
        \centering
        \captionsetup{justification=centering}
        \includegraphics[scale=.22]{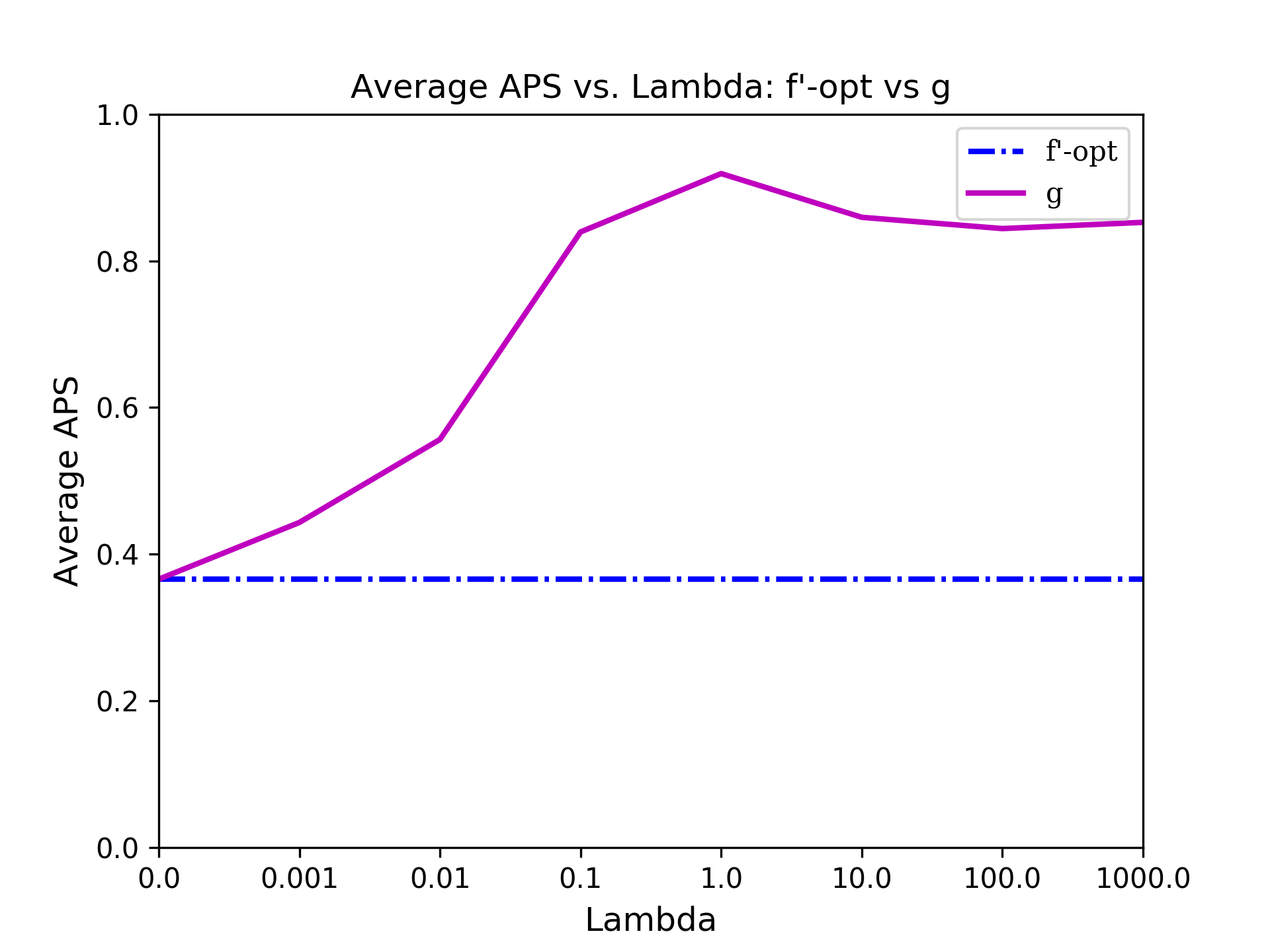}
        \caption{$B=4$.\label{fig:aps-benchb4}}
    \end{subfigure}
    \begin{subfigure}[]{.19\linewidth}
        \centering
        \captionsetup{justification=centering}
        \includegraphics[scale=.22]{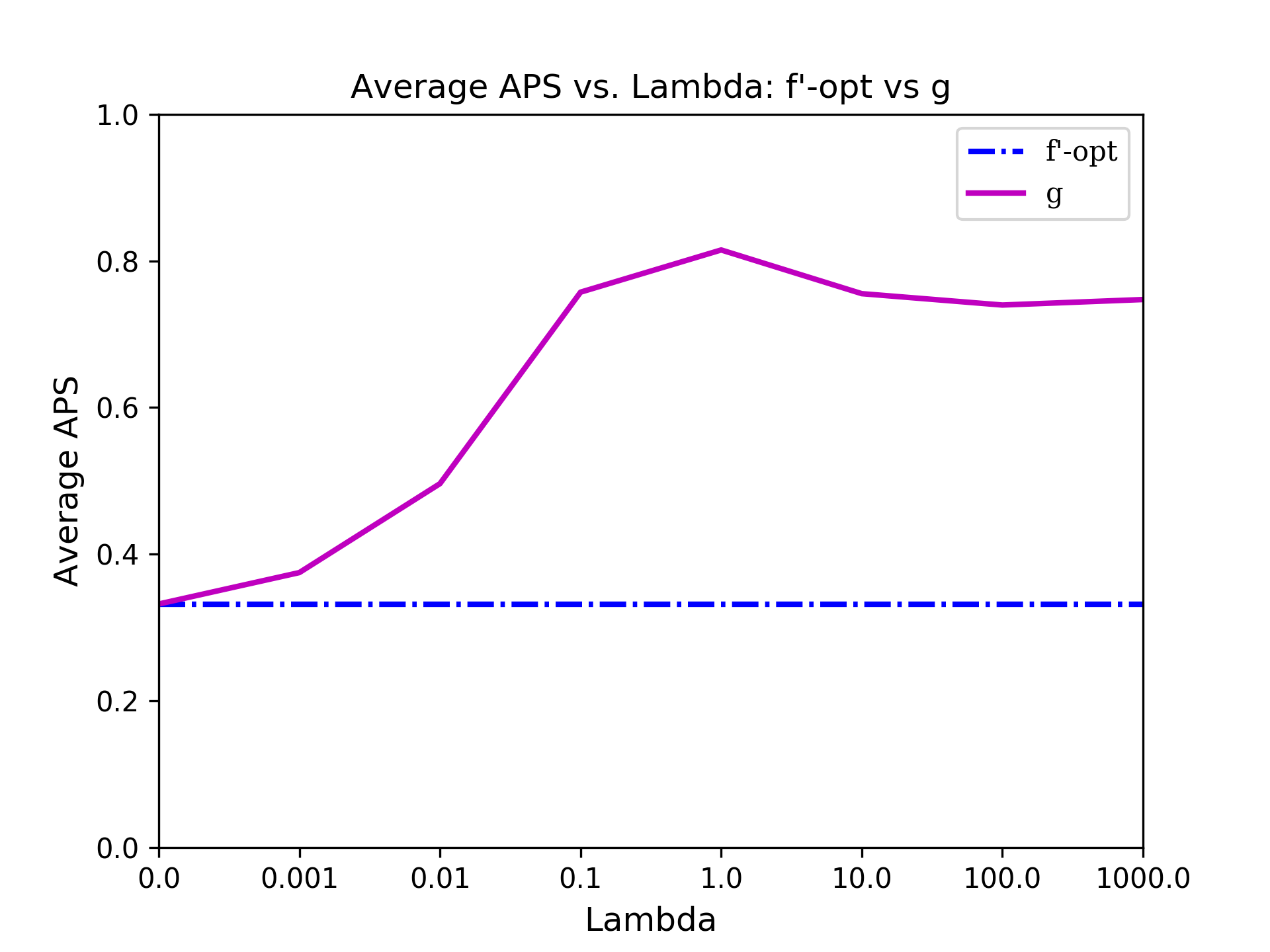}
        \caption{$B=6$.\label{fig:aps-benchb6}}
    \end{subfigure}
    \begin{subfigure}[]{.19\linewidth}
        \centering
        \captionsetup{justification=centering}
        \includegraphics[scale=.22]{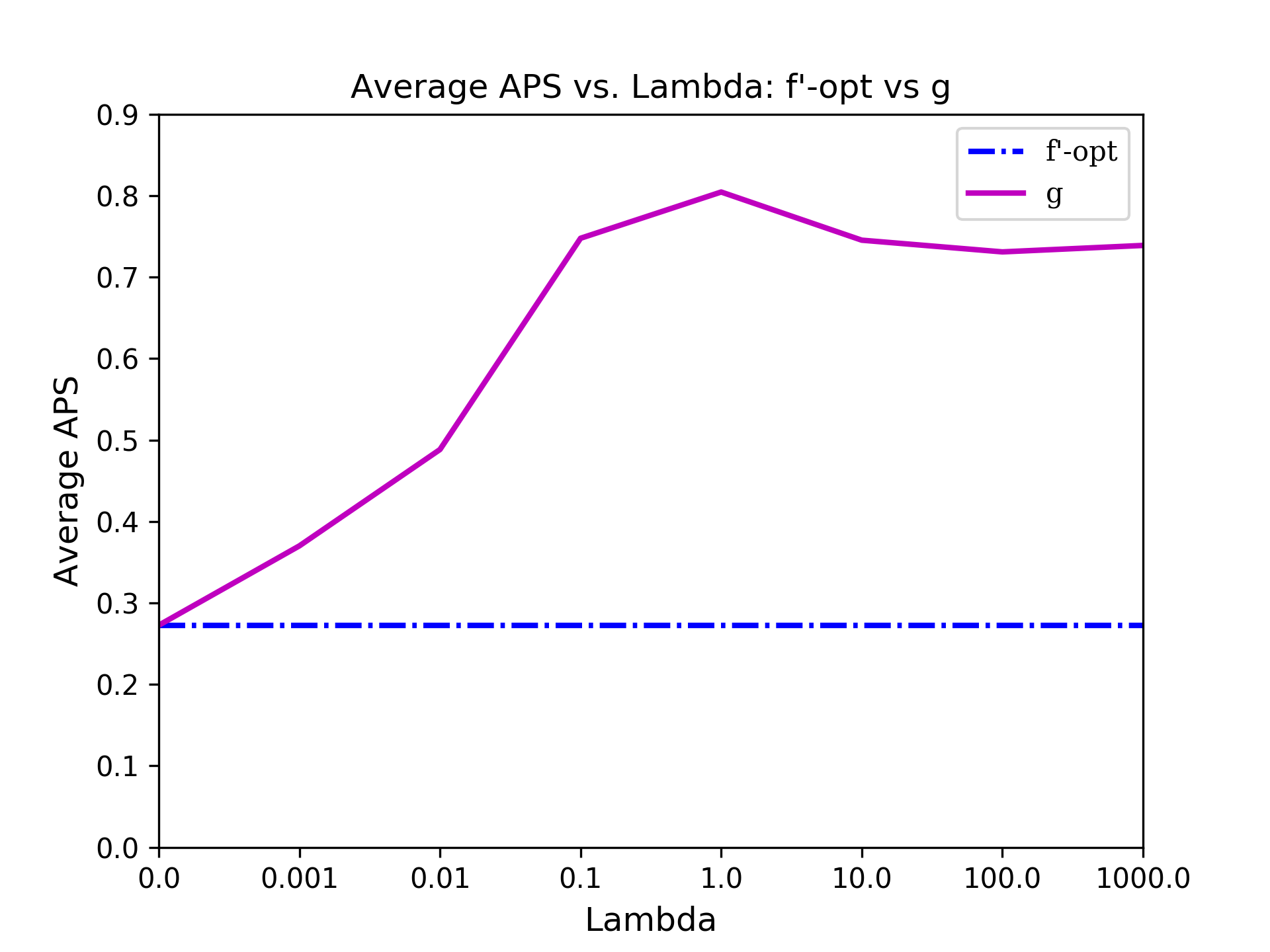}
        \caption{$B=8$.\label{fig:aps-benchb8}}
    \end{subfigure}
    \begin{subfigure}[]{.19\linewidth}
        \centering
        \captionsetup{justification=centering}
        \includegraphics[scale=.22]{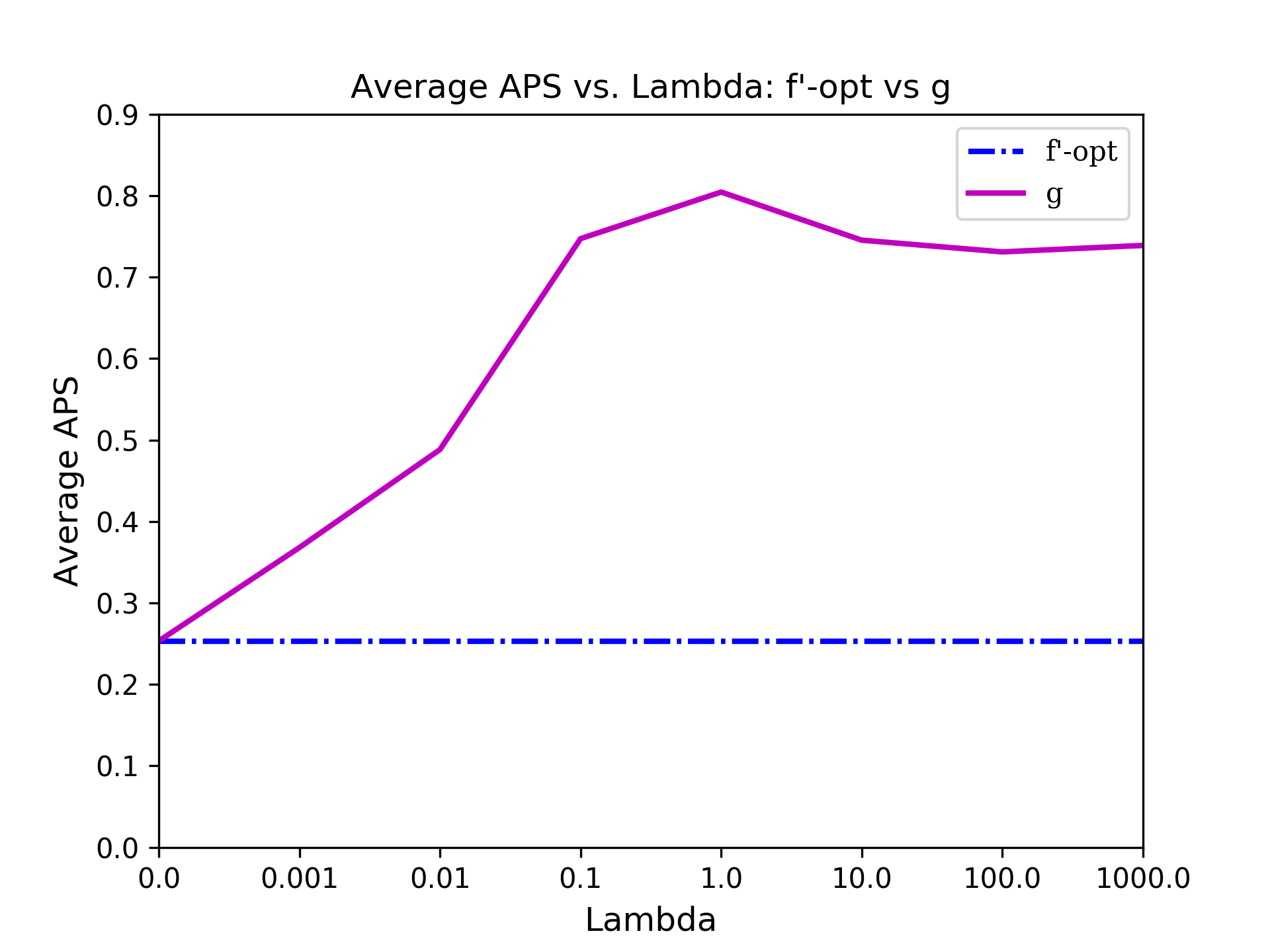}
        \caption{$B=10$.\label{fig:aps-benchb10}}
    \end{subfigure}
    \caption{$\lambda$ Calibration: Average $i$FEE and APS by Selected Budgetary Levels.\label{fig:lambda}}
\end{figure*}

In this subsection we examine the effect the $\lambda$ regularizer of $g$ has on average $i$FEE (Figures \ref{fig:lam-benchb2}-\ref{fig:lam-benchb10}); we compare the results to $f^{\prime}$-opt, also showing the corresponding average APS result (Figures \ref{fig:aps-benchb2}-\ref{fig:aps-benchb10}) to help inform the discussion. The results are presented in Figure \ref{fig:lambda} (note the different y-axis scales across the varying sub-figures).

There are several insights we can glean from examining Figure \ref{fig:lambda}. First, we can observe that the regularizer is functioning as designed: as we increase $\lambda$ the APS increases, although this effect seems to asymptote and even dip when $\lambda \geq 1$.  This effect is seen across the five budgetary levels presented. However, we also notice that while the average APS seems to asymptote there are further average $i$FEE improvements wrought from increasing $lambda$ beyond one. These $\lambda$-based $i$FEE improvements seem more apparent at lower to moderate budgetary levels (e.g., $B=4,6$). Second, we can see that, as in Figure \ref{fig:improve}, $g$ obtains better average $i$FEE scores at lower budgets, while at larger $B$, $f^{\prime}$-opt performs better (Figures \ref{fig:lam-benchb2}-\ref{fig:lam-benchb10}). Again, however, this result is tempered by that of the average APS, suggesting that the result of $f^{\prime}$-opt might not be entirely trustworthy at higher budgetary levels. Finally, in the aggregate, we can see that using $g$ one can get what one optimizes for: either $i$FEE improvement, causality (in terms of APS), or some combination thereof.

\subsubsection{Common recommendations}

We present the most common treatment policy recommendations by objective function for a budget $B=3$ in Table \ref{tab:common}. The values in this table were calculated by counting the number of instances that were recommended to adjust each respective feature. The most common treatment recommended by objective function are highlighted in \textcolor{red}{red}. 

\begin{table}[htp]
	\centering
	\begin{tabular}{lcccc}
	\toprule
    \textbf{Treatment/Obj} & $\mathbf{g(\pmb{\lambda}=10)}$ & \textbf{$\mathbf{f^{\prime}}$-opt} & \textbf{$\mathbf{f^{\prime}}$-no opt} & \textbf{$\mathbf{f}$}\\ \midrule\midrule
	Time out & \textcolor{red}{167} & 37 & 33 & 36 \\
	Daily Alco & 32 & 17 & 0 & 92\\
	Weekend Alco & 157 & 86 & 25 & 35 \\
	Absences & 90 & 75 & 12 & 88\\
    Study time & 129 & 124 & \textcolor{red}{306} & 74 \\
    Paid tutor & 127 & \textcolor{red}{266} & 1 & \textcolor{red}{193}\\\bottomrule
	\end{tabular}
	\caption{Most common treatments by objective function for $B=3$. \label{tab:common}}
\end{table}

First, there are clear differences in the most commonly recommended treatment, suggesting that our methodology has had an impact at the policy-level. Interestingly, the most commonly recommended treatment of $g$ (time out with friends) is the least commonly recommended treatment of $f^{\prime}$-opt and $f$; $f^{\prime}$-no opt has only 33 policies recommending this as well. Second, we notice that $g$ has relatively balanced recommendations among the different treatment options, which may suggest that what causes better grades for one person is different from that of another, which is not unexpected.

\section{Related Work}
In this section we briefly discuss past work related to this research. We stratify the discussion along two dimensions: \textbf{inverse classification} and \textbf{causal learning}.

\textbf{Inverse classification}-based works can be examined along numerous dimensions, including the framework/constraints, data type, and machine learning model capability; many of these components, which make up the inverse classification process, are inter-related and dependent upon one another (e.g., if the data are assumed to be discrete, then a certain set of optimization methodology should be used). Furthermore, not all inverse classification works are identified as such: action rules, which assume the data are discrete, is very comparable to inverse classification, but relies on rule-based methodology \cite{Tzacheva2017}.

Framework/constraints refer to the considerations made to the problem being modeled that produce real-world viable solutions (e.g., restricting the amount of change recommended). Our work builds upon recent past works that make the maximal number of considerations thusfar expressed in the literature \cite{Lash2017a,Lash2017b}. Other works make none \cite{Aggarwal2010,Pendharkar2002} or some \cite{Chi2012,Yang2012,Mannino2000,Barbella2009} of these considerations.

The data type simply refers to the type of data -- discrete \cite{Aggarwal2010,Chi2012,Yang2012,Tzacheva2017} or continuous \cite{Mannino2000,Barbella2009,Pendharkar2002,Lash2017a,Lash2017b}-- the underlying framework can support. In this work we use continuously valued data. 

Machine learning model capability refers to the versatility of the underlying framework w.r.t.~ the models it is capable of using.  Some inverse classification methods are constructed around specific classifiers (e.g., SVM) \cite{Aggarwal2010,Chi2012,Yang2012,Mannino2000,Barbella2009,Pendharkar2002}, while others permit the use of a variety of classification models \cite{Lash2017a,Lash2017b}. We adopt a framework that is capable of using differentiable classifiers, choosing to use neural networks. 

The classification model is closely linked to the optimization methodology, which is used to ``work backward'' through the model to elicit optimal treatment policies. These can be thought of as greedy (e.g., hill climbing) \cite{Aggarwal2010,Chi2012,Yang2012,Mannino2000} and non-greedy (e.g., gradient descent, branch and bound, etc.) \cite{Barbella2009,Lash2017a,Lash2017b,Pendharkar2002} methods. In this work we use the non-greedy projected gradient descent (PGD), as neural networks are differentiable and our constraints are linear.

There are many \textbf{causal learning}-based works and we, therefore, temper our discussion to those most relevant to this work and, when necessary, report only select papers. Causal works, specifically those focusing on inducing causal models using observational data, are typically framed around a single binary treatment \cite{Neyman1923,Rubin1974,Shalit2017,Johansson2016} or a single continuously-valued treatment \cite{Hirano2004}. In this work, we devised a method for inducing causal models having multiple, continuously-valued treatments.

The typical approach to causal modeling of observational data is through the potential outcomes framework (POF), otherwise referred to as the Neyman-Rubin framework (or, at times, just the Rubin Framework) \cite{Neyman1923,Rubin1974}. While the framework was originally formulated for a single, binary treatment, it has been updated to accommodate single, continuously-valued treatments as well \cite{Hirano2004}. In this paper, we argue that (a) because we are estimating the probability of a future outcome occurring and (b) because we are using multiple, continuously-valued treatments, the POF is unsuitable for our needs. We therefore update POF to ICPOF (inverse classification potential outcomes framework).  Additionally, the quantity of interest under the original POF is either the individual or average treatment effect (ITE/ATE). In ICPOF, we update the quantity of interest to $i$FEE (or individual feature-estimated effects).

Methods for eliciting causal effects from observational data include propensity score scaling \cite{Rosenbaum1983}, which includes inverse propensity score weighting \cite{Williamson2014,Lunceford2004} and covariate adjustment \cite{Pearl2009,Belloni2014,Shalit2017,Johansson2016},  propensity score matching \cite{Rosenbaum1989,Dehejia2002}, doubly robust estimation \cite{Bang2005,Funk2011}, and the use of instrumental variables \cite{Hernan2006,Baiocchi2014}. In this work, we expounded on inverse propensity score weighting methodology, which controls for observational data biases using the inverse of the propensity score. We develop the approximate propensity score weighting (APSW) method, based on Gaussian processes, to control for bias among our multiple, continuously-valued treatments that has immediate MLE connotations via reconstruction of the predictive distribution of each treatment at the individual (e.g., patient/student) level.

Causal learning, particularly in recent years, has been incorporated into machine learning/data mining-based methodology, and has also been found to be useful in a variety of application domains. Relevant methodology papers include neural network-based works \cite{Shalit2017,Johansson2016}, decision tree-based works \cite{Wager2017}, rule-based \cite{Wang2017}, and using Gaussian processes \cite{Alaa2017}, to name a few of many recent works. Furthermore, recent application-based innovation in the area of causal learning has been seen in the domains of logged bandit feedback \cite{Swaminathan2015} and advertising \cite{Schnabel2016,Hill2015}.

\section{Conclusions}

In this work we propose $Prophit$, a causal learning method and inverse classification framework for eliciting individual-specific treatment policies composed of multiple, continuously valued treatments. We accomplish this by developing APS, or approximate propensity score, that is based on Gaussian processes and subsequent reconstruction of the predictive distribution at the elicited unbiased treatment value estimate. Additionally, we update the well known potential outcomes framework (POF) to the inverse classification potential outcomes framework (ICPOF), which entails updating the equally well known individual treatment effect (ITE) score to the individual future estimate effects ($i$FEE), as well as developing a corresponding set of updated assumptions. We perform our experiments on student performance, a dataset freely available on the UCI Machine Learning repository. Subsequently, we conduct an in-depth analysis of our methodology. The result shows the validity and benefits of our method. Unfortunately, few freely available datasets are suitable for evaluating the methods disclosed in this work. Therefore, future work will involve the curation and analysis of suitable datasets capable of (a) capturing the considerations we make in this work and (b) evaluating methodology involving multiple, continuously valued treatments.

\bibliographystyle{ACM-Reference-Format}
\bibliography{causal_ic_paper}

\end{document}